%% file: main.tex
\documentclass[letterpaper,10pt,conference]{ieeeconf}

\IEEEoverridecommandlockouts  
                            
\usepackage{graphicx}
\usepackage{balance}
\usepackage{comment}
\usepackage{cite}
\usepackage{amssymb}
\usepackage[tight,footnotesize]{subfigure}
\usepackage[active]{srcltx}
\usepackage{amsmath}
\usepackage{mathtools}

\graphicspath{{./figs/}}
\renewcommand{\baselinestretch}{0.95}
\usepackage{eurosym}
\usepackage[plain]{algorithm}
\usepackage{algorithmic}
\usepackage{multicol}
\usepackage{dsfont}
\usepackage{amsfonts}

\usepackage{cases}
\usepackage{xcolor}

\begin{document}

\title{\LARGE \bf
Cooperative Receding Horizon 3D Coverage Control \\with a Team of Networked Aerial Agents
\vspace{-3mm}}

\author{Savvas~Papaioannou,~Panayiotis~Kolios,~Theocharis~Theocharides,\\~Christos~G.~Panayiotou and~Marios~M.~Polycarpou% <-this % stops a space
\thanks{The authors are with the KIOS Research and Innovation Centre of Excellence (KIOS CoE) and the Department of Electrical and Computer Engineering, University of Cyprus, Nicosia, 1678, Cyprus. E-mail:{\tt\small \{papaioannou.savvas, pkolios, ttheocharides, christosp, mpolycar\}@ucy.ac.cy}%
\newline
This work was undertaken as part of the GLIMPSE project EXCELLENCE/0421/0586 which is co-financed by the European Regional Development Fund and the Republic of Cyprus through the Research and Innovation Foundation's RESTART 2016-2020 Programme for Research, Technological Development and Innovation and supported by the European Union's Horizon 2020 research and innovation programme under grant agreement No 739551 (KIOS CoE), and from the Government of the Republic of Cyprus through the Cyprus Deputy Ministry of Research, Innovation and Digital Policy.
}}

\maketitle
\input{abstract}

\input{introduction}

\input{systemModel}

\input{approach}

\input{evaluation}

\input{conclusion}

\flushbottom
\balance

\bibliographystyle{IEEEtran}
\bibliography{IEEEabrv,main}

\end{document}

%% file: abstract.tex
\begin{abstract}

This work proposes a receding horizon coverage control approach which allows multiple autonomous aerial agents to work cooperatively in order cover the total surface area of a 3D object of interest. The cooperative coverage problem which is posed in this work as an optimal control problem, jointly optimizes the agents' kinematic and camera control inputs, while considering coupling constraints amongst the team of agents which aim at minimizing the duplication of work. To generate look-ahead coverage trajectories over a finite planning horizon, the proposed approach integrates visibility constraints into the proposed coverage controller in order to determine the visible part of the object with respect to the agents' future states. In particular, we show how non-linear and non-convex visibility determination constraints can be transformed into logical constraints which can easily be embedded into a mixed integer optimization program. 

%The effectiveness of the proposed approach is demonstrated through extensive simulation and real-world experiments.
\end{abstract}

%% file: introduction.tex
\section{Introduction} \label{sec:Introduction}
The recent technological advances in unmanned aerial vehicle (UAV) technology has led to a transformative impact in various application domains \cite{PapaioannouJ2,PapaioannouUnscented,Moraes2020,daud2022applications}. Specifically, during the last decade the problem of coverage path planning (CPP) \cite{Galceran2013} with UAVs has gained a lot of attention. Notably, the work in \cite{Li2011} proposes a coverage path planning method for a single UAV agent that optimizes the UAV's turning motion in planar convex polygonal areas, whereas in \cite{Xu2014} the authors consider a fixed-wing UAV, and utilize algorithms for the Chinese postman problem in order to compute an Eulerian path that covers all cells in the region of interest. An information theoretic terrain coverage planning approach is proposed in \cite{Liam2014} for a single fixed-wing UAV agent, whereas in \cite{Chen2021} a spatiotemporal clustering-based coverage approach is proposed for assigning regions of interest that need to be covered to a team of heterogeneous UAVs. The problem of multi-UAV CPP is also investigated in \cite{Theile2020} emphasizing on the energy efficiency of the mission. The multi-agent CPP problem for terrain coverage with workload balancing has recently been investigated in \cite{Collins2021}, and in \cite{Papaioannou2022,PapaioannouCybernetics} the coverage problem is investigated for 3D cuboid-like objects of interest. Despite the continuous advancements in this domain, there is still work to be done until this technology reaches the required level of maturity to enable autonomous coverage missions. The majority of UAV-based CPP approaches discussed above, and found in the literature, steer their focus towards covering mainly 2D planar areas and terrains\cite{PapaioannouTAES,PapaioannouCDC2023,Tan2021}, and not 3D objects. Moreover, coverage planning approaches which rely on simple geometric patterns (e.g., back-and-forth and zig-zag motions) \cite{Cabreira2018}, usually fail to generalize in 3D environments. Existing 3D coverage planning techniques often require specialized types of objects e.g., cuboids-like structures \cite{Papaioannou2022}, and mainly utilize UAVs equipped with fixed, and uncontrollable sensors \cite{Almadhoun2019} thus reducing the problem to a standard path-planning problem which does not accounts for the complexities of coverage control in 3D settings.

To tackle some of the challenges discussed above, in this work we propose a coverage controller which enables a team of autonomous UAV agents to compute cooperative finite-length look-ahead trajectories by jointly optimizing their kinematic and camera control inputs in order to cover in 3D the total surface area of an object of interest. In order to generate the look-ahead coverage plans we simulate the physical behavior of light in order to determine the visible parts of the object with respect to the future states of the agents, and we show how non-linear and non-convex visibility determination constraints can be transformed into logical constraints which can be embedded into a mixed integer optimization problem.

%In summary, the contributions of this work are the following:
%
%\begin{itemize}
%    \item We formulate the problem of cooperative coverage control in 3D environments as an optimal control problem, which jointly optimizes the kinematic and camera control inputs of a team of UAV agents, and allows the total surface area of an object of interest to be covered, while considering coupling constraints amongst the team of agents which discourage the duplication of work.
%    \item In order to generate look-ahead coverage trajectories we emulate the physical behavior of light to determine the visible parts of the object with respect to the future states of the agents. In particular we show how non-linear and non-convex visibility determination constraints can be transformed into logical constraints and embedded into a mixed integer optimization problem, which can be solved using off-the-shelf tools.
%    \item Finally, extensive simulation and real-world experiments demonstrate the performance of the proposed approach.
%\end{itemize}

The rest of the paper is organized as follows. Section \ref{sec:system_model} develops the system model, Section \ref{sec:problem} formulates the problem tackled, and Section \ref{sec:approach} discusses the details of the proposed approach. Finally, Section \ref{sec:Evaluation} evaluates the proposed approach, and Section \ref{sec:conclusion} concludes the paper, and discusses future directions.

%% file: systemModel.tex
\section{Preliminaries} \label{sec:system_model}

\subsection{Agent Kinematic Model} \label{ssec:kinematic_model}

We assume that a cooperative team of $N$ autonomous networked aerial agents denoted by $j \in [1,..,N]$, operate inside a bounded 3D environment $\mathcal{E} \subset \mathbb{R}^3$, with discrete-time kinematics given by the following linear state-space model:
\begin{equation}\label{eq:kinematics}
x^j_{k} = 
\begin{bmatrix}
    \boldsymbol{1}_{3\times3} & \Delta T \times \boldsymbol{1}_{3\times3}\\
    \boldsymbol{0}_{3\times3} & (1-\gamma) \times \boldsymbol{1}_{3\times3}
   \end{bmatrix} x^j_{k-1} + 
\begin{bmatrix}
    \boldsymbol{0}_{3\times3} \\
     \frac{\Delta T}{m} \times \boldsymbol{1}_{3\times3}
   \end{bmatrix} u^j_{k},
\end{equation}

\noindent which is abbreviated hereafter as $x^j_{k}~=~Ax^j_{k-1}~+~Bu^j_{k}$, where $x^j_k = [(x_k^{j,\mathbf{p}})^\top,(x_k^{j,\mathbf{v}})^\top]^\top \in \mathcal{X} \subset  \mathbb{R}^6$ denotes agent's $j$ kinematic state at time-step $k$, which is composed of position (i.e., $x_k^{j,\mathbf{p}} \in \mathbb{R}^3 $), and velocity (i.e., $x_k^{j,\mathbf{v}} \in \mathbb{R}^3$) components, in the 3D cartesian coordinate system. It is assumed that the agents are controllable, and that can be commanded to execute a certain direction and speed through the control input $u^j_k \in \mathcal{U} \subset \mathbb{R}^3, j \in  [1,..,N]$ which denotes the applied input force. In Eq. \eqref{eq:kinematics} the parameter $\Delta T$ denotes the sampling interval, $\gamma$ models the air resistance coefficient, and finally $m$ is agent's $j$ mass, which without loss of generality is assumed to be the same for all agents. In this work it is assumed that all agents maintain a wireless communication link with a central mobile base-station which is used for information exchange.

%The matrices  $\boldsymbol{1}_{3\times3}$ and $\boldsymbol{0}_{3\times3}$ denote the 3-by-3 identity and zero matrices respectively.
%i.e., sending mission data and receiving cooperative control commands. 

%The resulting coverage trajectory can be used as the reference trajectory of a low-level controller (i.e., an autopilot) \cite{Simone2020,Elkaim2015,Garcia2011,Yang2016}.

%This assumption is made here merely to simplify the platform requirements for testing and the evaluation of the proposed controller in real-world settings. However, as we discuss in Sec. \ref{sec:approach} the proposed controller can be easily formulated as a distributed system.
%We should also mention here that although Eq. \eqref{eq:kinematics} does not describes the true underline aerodynamical behavior of the robot (i.e., the UAV), it can be used  to construct the desired mission trajectory, which in turn can be tracked with the appropriate (i.e., depending on the robot type such as rotor-copter or fixed-wing UAV) low-level guidance and navigation controller (e.g., an auto-pilot) \cite{Simone2020,Elkaim2015,Garcia2011,Yang2016}.

\subsection{Agent Camera Model} \label{ssec:sensing_model}
Each agent $j \in [1,..,N]$ is equipped with a camera system attached to a gimbal device which allows the 3D rotation of the camera's finite field-of-view (FOV). In this work, the camera's FOV is modelled as a regular right pyramid which exhibits four triangular lateral faces and a rectangular base. The camera optical center is positioned directly above the centroid of the rectangular FOV base. In essence, the camera's FOV is determined in this work by the parameter set $(\ell,w,r)$, where $\ell$ and $w$ are the length and width of the FOV rectangular base respectively, and $r$ (i.e., the height of the pyramid) determines the FOV range. 
Let the five FOV vertices of a downward facing camera, centered at the origin of the 3D cartesian coordinate system to be given by the 3-by-5 matrix $\mathcal{V}_0$ as:
\begin{equation}
    \mathcal{V}_0 =
    \begin{bmatrix}
       -\frac{\ell}{2} & \frac{\ell}{2} & \frac{\ell}{2}  & -\frac{\ell}{2} & 0 \\
        \frac{w}{2} &\frac{w}{2} & -\frac{w}{2} &  -\frac{w}{2} & 0 \\
        -r  & -r  &  -r  &  -r  & 0 \\
    \end{bmatrix}.
\end{equation}

\noindent The FOV is rotated in 3D space by commanding the gimbal controller to execute sequentially two elemental rotations i.e., one rotation by angle $\theta \in [0,\pi)$ around the $y-$axis, followed by a rotation $\phi \in [0,2\pi)$ around the $z-$axis. Therefore, at each time-step $k$ the agent $j$ with position $x_k^{j,\mathbf{p}}$ can rotate the camera's FOV anywhere inside the surveillance region, via the following geometric transformation:
\begin{equation}\label{eq:fov_eq}
  \!\! \mathcal{V}^j_k(\theta^j_k,\phi^j_k,x_k^{j,\mathbf{p}})^i \!=\! R_{z}(\phi^j_k) R_{y}(\theta^j_k) \mathcal{V}^i_0 \!+\!  x_k^{j,\mathbf{p}},\forall i\!\in\![1,..,5]
\end{equation}
\noindent where $\mathcal{V}^i_0$ denotes the $i_\text{th}$ column of $\mathcal{V}_0$, and therefore  $\mathcal{V}^j_k(\theta^j_k,\phi^j_k,x_k^{j,\mathbf{p}})^i$ is the corresponding rotated and translated vertex of the FOV. The parameters $\theta^j_k$ and $\phi^j_k$ are the input rotation angles, and $R_{y}(\alpha)$, $R_{z}(\alpha)$ represent the basic 3-by-3 rotation matrices \cite{Taubin2011} which rotate vectors by an angle $\alpha$ around the $y-$ and $z-$axis respectively.
%\begin{equation}
%	\begin{bmatrix}
%       \text{cos}(\alpha) & 0 & \text{sin}(\alpha)\\
%       0 & 1 & 0\\
%       -\text{sin}(\alpha) & 0 & \text{cos}(\alpha)    
%    \end{bmatrix}, ~ \begin{bmatrix}
%       \text{cos}(\alpha) & -\text{sin}(\alpha) & 0\\
%       \text{sin}(\alpha) &  \text{cos}(\alpha) & 0\\
%        0 & 0 & 1
%    \end{bmatrix}.
%\end{equation}
%\begin{align}
% R_{\theta}(\alpha) &=
%    \begin{bmatrix}
%       \text{cos}(\alpha) & 0 & \text{sin}(\alpha)\\
%       0 & 1 & 0\\
%       -\text{sin}(\alpha) & 0 & \text{cos}(\alpha)    
%    \end{bmatrix},\\
%    R_{\phi}(\alpha) &=
%    \begin{bmatrix}
%       \text{cos}(\alpha) & -\text{sin}(\alpha) & 0\\
%       \text{sin}(\alpha) &  \text{cos}(\alpha) & 0\\
%        0 & 0 & 1
%    \end{bmatrix}.
%\end{align}
%In the discussion above, in order to simplify the analysis, and without loss of generality we have assumed that the camera coordinate frame and the world coordinate frame are aligned and described by a regular right-handed 3D cartesian coordinate system. Therefore, Eq. \eqref{eq:fov_eq} ultimately describes the camera's FOV pose (position and orientation), with the camera's optical center (i.e., the FOV apex) to be given by the agent's position $x_k^{\mathbf{p}}$. 
We assume that the gimbal device is bounded to operate within a predefined finite set of admissible input rotation angles $\Xi = \Theta \times \Phi = \{(\theta,\phi) | \theta \in \Theta, \phi \in \Phi\}$, where $\times$ denotes the Cartesian product on the finite sets $\Theta$ and $\Phi$. Therefore, the camera FOV of each agent $j$ can take, at each time-step $k$, one out of $|\Xi|$ possible configurations ($|\Xi|$ denotes the cardinality of the set $\Xi$). 

Finally, it is assumed that at each time-step $k$ a finite set of (straight) light-rays, which model the direction of the propagation of light, enter the camera's optical center and cause matter to be imaged. The set of light-rays captured through the agent's camera FOV $\mathcal{V}^j_k(\theta^j_k,\phi^j_k,x_k^{j,\mathbf{p}})$ is denoted in this work as $\mathcal{L}^j_k(\theta^j_k,\phi^j_k,x_k^{j,\mathbf{p}}) = \{\Lambda^j_{k,1},..,\Lambda^j_{k,n}\}$, where $\Lambda^j_{k,i}$ denotes the individual light-ray in the set which is further given by the line-segment:
\begin{equation}\label{eq:lightray}
    \Lambda^j_{k,i} = \lambda^j_{k,i} + d(x_k^{j,\mathbf{p}}-\lambda^j_{k,i}), ~\forall d \in [0,1],
\end{equation}

\noindent 
where $x_k^{j,\mathbf{p}}$ is the light-ray's end point which enters the camera's optical center at time-step $k$, $\lambda^j_{k,i}$ is a fixed point on the camera's FOV base denoting the ray's origin, and $d$ is a scalar. Note here that every FOV state $\mathcal{V}^j_k(\theta^j_k,\phi^j_k,x_k^{j,\mathbf{p}})$ generates a different set of light-rays $\mathcal{L}^j_k(\theta^j_k,\phi^j_k,x_k^{j,\mathbf{p}})$.

The goal of the agents is to cover with their cameras the total surface area $\partial \mathcal{O}$ of a known object of interest $\mathcal{O} \in \mathcal{E}$. This object's surface area has been triangulated into a finite set of non-overlapping triangular facets $\tau \in \mathcal{T}$, where $\mathcal{T}$ is the object's surface triangle mesh. Consequently, our aim becomes the generation of cooperative coverage trajectories which  cover all facets $\tau \in \mathcal{T}$.

%\subsection{Object of Interest Model}\label{ssec:object_of_interest}
%
%
%The goal of the agents is to cover or observe with their cameras the total surface area $\partial \mathcal{O}$ of an object of interest $\mathcal{O} \in \mathcal{E}$ which is assumed to be known prior to the mission. In essence, the object of interest is 3D reconstructed as a point-cloud from multiple calibrated images that have been collected prior to the mission. Let us denote the 3D point-cloud representation of the object's surface area as $\mathcal{P} = \{p_1,..,p_{|\mathcal{P}|}\} \subset \partial \mathcal{O}$. We use Delaunay triangulation in order to triangulate the set of points in $\mathcal{P}$, and generate a triangle mesh $\mathcal{T}$ which decomposes the object's surface area into a finite set of non-overlapping triangular facets $\tau \in \mathcal{T}$. Consequently, our aim becomes the generation of cooperative coverage trajectories which  cover all facets $\tau \in \mathcal{T}$.

% as shown in Fig. \ref{fig:fig1}. 

%\begin{figure}
%	\centering
%	\includegraphics[width=\columnwidth]{figs/fig1.pdf}
%	\caption{The surface area $\partial \mathcal{O}$ of the object of interest $\mathcal{O}$ is 3D reconstructed as a point-cloud $\mathcal{P}$, which is then triangulated to generate the triangle mesh $\mathcal{T}$ as shown above.}	
%	\label{fig:fig1}
%	\vspace{-0mm}
%\end{figure}

\section{Problem Formulation} \label{sec:problem}
\textit{
Given a team of $N$ cooperative agents $j = [1,..,N]$, find the joint kinematic (i.e., input force $u^j_k, \forall j$) and camera (i.e., rotation angles $(\theta^j_k, \phi^j_k), \forall j$) control inputs over a sufficiently large planning horizon of length $K^\prime$ time-steps which result in the optimal coverage of the total surface area $\mathcal{T}$ of the object of interest $\mathcal{O}$}.
A high-level formulation of the coverage problem discussed above is shown in Problem (P1), posed as an optimal control problem. As shown in Problem (P1), we are seeking to find the agents' joint control inputs $\{u^j_k,~\theta^j_k,~\phi^j_k\}_{j=1}^N$ over a finite planning horizon $k = [1,..,K^\prime]$ of length $K^\prime$ time-steps which optimize a mission related objective function (e.g., the coverage elapsed time) denoted as $\mathcal{F}_\text{mission}$, and shown in Eq. \eqref{eq:objective_P1}, subject to the set of coverage constraints shown in Eq. \eqref{eq:P1_1}-\eqref{eq:P1_2}.

\begin{algorithm}
\vspace{-5mm}
\begin{subequations}
\begin{align}
&\hspace*{-13mm}\textbf{(P1)}~\texttt{Coverage Problem~~~~~~} & \hspace*{-25mm}  \nonumber\\
& \hspace*{-13mm}~~~\underset{\{u^j_k,~\theta^j_k,~\phi^j_k\}_{j=1}^N}{\arg \min} ~\mathcal{F}_\text{mission}& \label{eq:objective_P1} \\
&\hspace*{-13mm}\textbf{subject to: $k \in [1,..,K^\prime]$, $j \in [1,..,N]$  } ~  & \nonumber\\
&\hspace*{-13mm}  x^j_{k} = A^{k} x^j_{0} + \sum_{\kappa=1}^{k} A^{k-\kappa} B u^j_\kappa & \hspace*{-25mm} \forall k, \forall j \label{eq:P1_1}\\
&\hspace*{-13mm} \exists \kappa \leq K^\prime : \tau \in \bigcup_{j=1}^N \bigtriangleup \left(\mathcal{V}^j_{\kappa}(\theta^j_{\kappa},\phi^j_{\kappa},x_{\kappa}^{j,\mathbf{p}}) \right) & \hspace*{-25mm} \forall \tau \in \mathcal{T}  \label{eq:P1_2}
%&\hspace*{-7mm} \sum_{j=1}^N g^j_{c,k} \leq 0 & \hspace*{-25mm}  \forall k, \forall c \in C \label{eq:P1_4}\\
%&\hspace*{-7mm} x^{j,\mathbf{p}}_0, x^{j,\mathbf{p}}_k \notin \triangle(\psi) & \hspace*{-25mm}  \forall \psi \in \Psi, \forall k \label{eq:P1_5}\\
%&\hspace*{-7mm} x^j_0, x^j_{k} \in \mathcal{X}, ~ u^j_k \in \mathcal{U}, \theta^j_k \in \Theta,~\phi^j_k \in \Phi& \hspace*{-25mm}  \forall k,\forall j\label{eq:P1_6}
\end{align}
\end{subequations}
\vspace{-6mm}
\end{algorithm}

The constraint in Eq. \eqref{eq:P1_1} is due to the agent's kinematic model as discussed in Sec. \ref{ssec:kinematic_model} which has been obtained from the recursive application of Eq. \eqref{eq:kinematics} for time-steps $k=[1,..,K^\prime]$, with a known initial state $x^j_0 = \bar{x}^j$.
The next constraint shown in Eq. \eqref{eq:P1_2} ensures that the total surface area $\mathcal{T}$ of the object of interest is cooperatively covered by the agents during the mission. In order to achieve this we require that for each facet $\tau \in \mathcal{T}$ there exists a time-step $\kappa \leq K^\prime$ for which $\tau$ resides within the convex-hull (indicated by the $\triangle$ operator) of some agent's camera FOV. In essence we require that each facet $\tau$ is covered by some agent $j$. However, in order to be able plan accurate look-ahead coverage trajectories, we first need to determine which parts of the object are actually visible given the future states of the agents (i.e., determine the visible area given the predicted camera poses). In that sense, the constraint in Eq. \eqref{eq:P1_2} is actually incomplete since it does not indicate what is the observable FOV. In the next section we show how we have incorporated visibility determination constraints into Problem (P1), in an effort to generate cooperative look-ahead coverage plans. 

%We will also show how Problem (P1) can be solved more efficiently by reformulating it into a receding horizon optimal control problem, and how coupling constraints amongst the agents can be used to improve performance by minimizing the duplication of work. 

%% file: approach.tex
\section{Cooperative Receding Horizon 3D Coverage Control}\label{sec:approach}

The coverage planning problem discussed in the previous section is quite challenging to be solved efficiently. In particular, observe that a feasible solution to this problem is directly coupled with the length of the planning horizon $K^\prime$ i.e., if $K^\prime$ is too short, then no feasible solution may exist, while if $K^\prime$ is too long then the computational complexity increases unnecessarily.
For the reasons discussed above, Problem (P1) is re-formulated as a receding horizon optimal control problem in where the joint control inputs over all agents $j=[1,..,N]$ are computed in an on-line fashion at each time-step $k$ inside a rolling finite planning horizon $K$ i.e., $\{u^j_{k+\kappa|k},~\theta^j_{k+\kappa|k},~\phi^j_{k+\kappa|k}\}, \forall \kappa \in [1,..,K]$. As a result, at each time-step $k$ the agents plan finite-length look-ahead coverage trajectories $x^j_{k+\kappa|k}, \kappa = [1,..,K]$, where the notation $x_{\kappa|k}$ denotes the predicted agent state at time-step $\kappa \geq k$, which was computed at time-step $k$.

\begin{figure}
	\centering
	\includegraphics[width=\columnwidth]{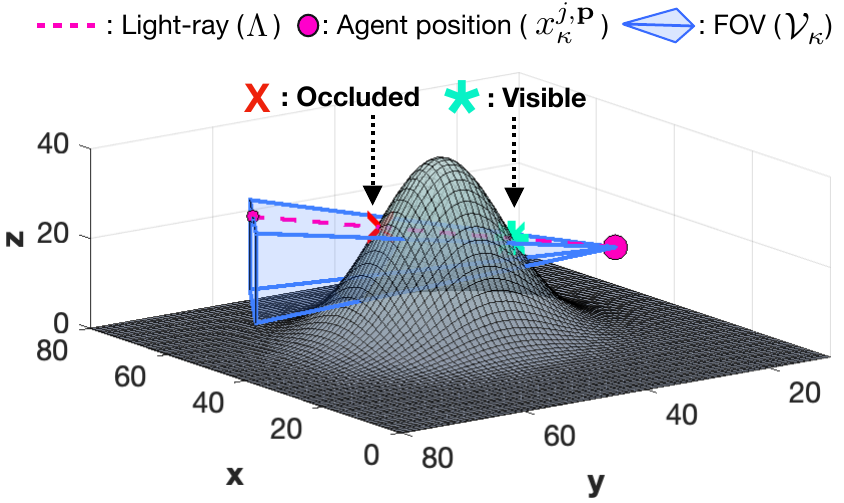}
	\caption{Illustrative example of the visibility determination problem, where $x_\kappa^{j,\mathbf{p}}$ is the future predicted agent position at time-step $k+\kappa|k$, and $\mathcal{V}_\kappa$ is the predicted FOV configuration at the same time-step. With the FOV pose shown in the figure the facet indicated with $\times$ is not visible as there is no light-ray that can be traced back to it i.e., all light-rays are blocked. On the other hand the facet indicated with a $\star$ can be observed as shown.}	
	\label{fig:fig1}
	\vspace{-6mm}
\end{figure}

\subsection{Visibility Determination}\label{ssec:visibility_determination}
In order to generate the agents' future coverage trajectories over a finite planning horizon $K$, we first need to have a way of determining which parts of the object's surface area are visible given the future planned states of the agents at time-steps $k+\kappa|k, \kappa \in [1,..,K]$, abbreviated for simplicity as $\kappa$.
Intuitively, the facet $\tau \in \mathcal{T}$, is visible through the agent's $j$ camera FOV $\mathcal{V}^j_{\kappa}(\theta^j_{\kappa},\phi^j_{\kappa},x_\kappa^{j,\mathbf{p}})$ at time-step $\kappa$ when: a) $\tau$ resides inside the convex-hull of the agent's camera FOV i.e., $\tau \in \triangle \left(\mathcal{V}^j_\kappa(\theta^j_\kappa,\phi^j_\kappa,x_\kappa^{j,\mathbf{p}}) \right)$, and b) there exists a light-ray $\Lambda^j_{\kappa,i} \in \mathcal{L}^j_\kappa(\theta^j_\kappa,\phi^j_\kappa,x_\kappa^{j,\mathbf{p}})$ which enters the camera's optical center and can be traced back to facet $\tau$. On the other hand, when no light-ray can be traced back to $\tau$, indicates that the specific facet is not visible since the propagation of light is blocked. Specifically, the notion of visibility can now be defined as follows: The facet $\tau \in \mathcal{T}$ is visible through the agent's $j$ camera FOV $\mathcal{V}^j_\kappa(\theta^j_\kappa,\phi^j_\kappa,x_\kappa^{j,\mathbf{p}})$ at the future time-step $\kappa$ when:
\begin{equation}\label{eq:visibility_eq}
\exists \Lambda^j_{\kappa,i} \in \mathcal{L}^j_\kappa(\theta^j_\kappa,\phi^j_\kappa,x_\kappa^{j,\mathbf{p}}) : \Lambda^j_{\kappa,i} \oplus \mathcal{T} = \tau,
\end{equation}
\noindent where the operator $\oplus$ returns the facet $\tau \in \mathcal{T}$ which intersects last with the light-ray $\Lambda^j_{\kappa,i}$; otherwise it returns $\emptyset$ if no facet $\tau$ intersects with the light-ray. 
Let us denote the equation of the plane which contains facet $\tau$ as $\alpha_{\hat{\tau}} \cdot x = \beta_{\hat{\tau}}$, where $\hat{\tau} \in [1,..,|\mathcal{T}|]$ is the index pointing to facet $\tau \in \mathcal{T}$, $\alpha_{\hat{\tau}} \in \mathbb{R}^3$ is the unit outward normal vector to the plane containing $\tau$, $x \in \mathbb{R}^3$, $\beta_{\hat{\tau}}$ is a scalar, and the notation $a \cdot  b$ denotes the dot product of the vectors $a$ and $b$. Subsequently, the operation $\Lambda^j_{\kappa,i} \oplus \tau$ finds the intersection point (if exists) between the light-ray $\Lambda^j_{\kappa,i}$ which is given by Eq. \eqref{eq:lightray}, and the plane which contains facet $\tau$ as follows:
\begin{subequations}
\begin{align}
    &\alpha_{\hat{\tau}} \cdot [\lambda^j_{\kappa,i} + d(x_\kappa^{j,\mathbf{p}}-\lambda^j_{\kappa,i})] = \beta_{\hat{\tau}} \implies \label{eq:r1}\\
    &d = \frac{\beta_{\hat{\tau}} - \alpha_{\hat{\tau}} \cdot \lambda^j_{\kappa,i}}{\alpha_{\hat{\tau}} \cdot (x_\kappa^{j,\mathbf{p}}-\lambda^j_{\kappa,i})} \label{eq:r2},
\end{align}
\end{subequations}
\noindent where Eq. \eqref{eq:r1} is the result of the substitution of $\Lambda^j_{\kappa,i}$ for $x$ in the equation of the plane which contains facet $\tau$, and then in Eq. \eqref{eq:r2} we solve for $d$.  Consequently, if the denominator of Eq. \eqref{eq:r2} is equal to zero, the light-ray $\Lambda^j_{\kappa,i}$ and the facet $\tau$ are parallel which results in either no visibility (i.e., $\beta_{\hat{\tau}} - \alpha_{\hat{\tau}} \cdot \lambda^j_{\kappa,i} \neq 0 $) or distorted view (i.e., when $\beta_{\hat{\tau}} - \alpha_{\hat{\tau}} \cdot \lambda^j_{\kappa,i} = 0 $). In essence, we are interested in the scenario where there exists a single point intersection between the light-ray $\lambda^j_{\kappa,i}$ and the plane which contains the facet $\tau$ i.e., when $\alpha_{\hat{\tau}} \cdot (x_\kappa^{j,\mathbf{p}}-\lambda^j_{\kappa,i}) \ne 0$ and $d \in [0,1]$ which indicates the visibility of $\tau$ by the light-ray $\lambda^j_{\kappa,i}$ i.e., the light-ray is not blocked and traces back to $\tau$, as illustrated in Fig. \ref{fig:fig1}.

The procedure discussed above must be evaluated at each time-step inside the planning horizon for all pairwise combinations of light-rays $\lambda^j_{\kappa,i} \in \mathcal{L}^j_\kappa(\theta^j_\kappa,\phi^j_\kappa,x_\kappa^{j,\mathbf{p}})$, and facets $\tau \in \mathcal{T}$ which not only is computationally expensive, but also requires the integration of non-convex and non-linear constraints which are challenging to be handled efficiently during optimization. In order to bypass this challenge, in this work we follow an alternative procedure which allows us to first learn a set of state-dependant constraints for determining visibility, and subsequently incorporating these constraints into the coverage planning optimization problem as logical constraints.

To achieve this the environment $\mathcal{E}$ is first decomposed into a 3D grid $\tilde{\mathcal{E}}$ consisting of a finite number of disjoint cells i.e., $\tilde{\mathcal{E}}=\{c_1,..,c_{|\tilde{\mathcal{E}}|}\}$, such that $\bigcup_{i=1}^{|\tilde{\mathcal{E}}|} c_i= \tilde{\mathcal{E}}$. Then, within each cell $c \in \tilde{\mathcal{E}}$, we randomly sample the agent states i.e., $(x^{\mathbf{p}},\theta,\phi)$ and then the visibility determination procedure discussed above i.e., Eq. \eqref{eq:r2} is performed for various configurations of the FOV state $\mathcal{V}(\theta,\phi,x^{\mathbf{p}})$ to identify the visible facets $\tau$. Let us denote with $\tilde{\mathcal{L}}_{\hat{c}}$ the set of light-rays that have been obtained from the application of $N_r$ random joint combinations of the parameter set $(x^{\mathbf{p}}, \theta, \phi)_i, ~i =[1,..,N_r]$ inside cell $c$ i.e., $ \tilde{\mathcal{L}}_{\hat{c}} = \bigcup_{i=1}^{N_r} \{\Lambda : \Lambda \in  \mathcal{L}^i_{\hat{c}}(\theta_i,\phi_i,x_i^{\mathbf{p}})\}$,  where $\hat{c} \in [1,..,|\tilde{\mathcal{E}}|]$ is the index pointing to cell $c \in \tilde{\mathcal{E}}$, $\mathcal{L}^i_{\hat{c}}(\theta_i,\phi_i,x_i^{\mathbf{p}})$ is the set of light-rays given by the camera pose obtained with the $i_\text{th}$ set of control inputs $(\theta_i,\phi_i)$, for the agent location $x_i^{\mathbf{p}}$ sampled within cell $c$. We then learn the following logical visibility determination constraints: 
\begin{equation} \label{eq:vis_con1}
    \rho_{\hat{c},{\hat{\tau}}} = 1 \iff \exists \Lambda \in \tilde{\mathcal{L}}_{\hat{c}} : \Lambda \oplus \mathcal{T} = \tau,~\forall \hat{c},\hat{\tau}
\end{equation}
\noindent Once the constraints above are learned, we can determine the visibility of facet $\tau$, by utilizing the binary variable $\rho_{\hat{c},\hat{\tau}}$ which is activated when there exists a light-ray $\Lambda$ which traces back to facet $\tau$ when the agent is within the cell $c$.

\subsection{Coverage Objective Function} \label{ssec:objective}
Let us assume that each facet $\tau \in \mathcal{T}$ can be uniquely identified by the agent $x^j_{k+\kappa|k}$, and thus its state (i.e., observed/covered or unobserved) can be associated with the binary variable $\hat{b}^j_{\hat{\tau},\hat{\xi},k+\kappa|k} \in \{0,1\}$ which indicates that facet $\tau$ (indexed by $\hat{\tau} \in [1,..,|T|]$), is planned to be observed and covered by agent $j$, with the FOV rotation $\xi \in \Xi$ (indexed by $\hat{\xi} \in [1,..,|\Xi|]$) at the future time-step $k+\kappa|k, \kappa \in [1,..,K]$ of the planning horizon (abbreviated simply as $\kappa$ throughout the paper). As a reminder the FOV rotaion $\xi \in \Xi$ maps to a certain set of rotation angles $(\theta,\phi)$ i.e., $\xi = (\theta,\phi) \in \Theta \times \Phi$ as discussed in Sec. \ref{ssec:sensing_model}.
The cooperative coverage objective function $\mathcal{F}_\text{mission}$ to be optimized over the planning horizon $k+\kappa|k, \kappa \in [1,..,K]$, shown in Eq. \eqref{eq:objective_P2}, can now be defined as $\mathcal{F}_\text{mission} = \sum_{j=1}^{N} F^j$, where $F^j$ is agent's $j$ own coverage objective function which is further given by:
\begin{equation}\label{eq:objective_agent}
   F^j = \mathcal{D}(x_{k+1|k}^{j,\mathbf{p}},\tau^j_\star) - \sum_{\kappa=1}^K \sigma(\kappa) \sum_{\hat{\xi}=1}^{|\Xi|} \sum_{\hat{\tau}=1}^{|\mathcal{T}|} \hat{b}^j_{\hat{\tau},\hat{\xi},k+\kappa|k} ,
\end{equation}
\noindent which essentially is minimized when all facets $\tau \in \mathcal{T}$ are planned to be covered inside the planning horizon i.e., $\hat{b}^j_{\hat{\tau},\hat{\xi},k+\kappa|k} =1, \forall \hat{\tau}$, assuming a sufficiently large enough value of $K$. Otherwise, $F^j$ incentivizes the agent to generate a trajectory which covers as many facets as possible inside the planning horizon. The time-dependent term $\sigma(\kappa)$ penalizes facets that are covered later in the horizon i.e., $\sigma(\kappa) = K-(\kappa-1)$ or any other penalty scheme can be used. Finally, the term $\mathcal{D}(x_{k+1|k}^{j,\mathbf{p}},\tau^j_\star)=||x_{k+1|k}^{j,\mathbf{p}}-\tau^j_\star||^2_2$ drives the agent $j$ towards the centroid of its nearest unobserved facet $\tau^j_\star$, in order to make sure that the mission can progress particularly in the events where no facets can be reached for coverage inside the planning horizon.

\begin{algorithm}
\begin{subequations}
\begin{align} 
&\hspace*{-3mm}\textbf{(P2)}~\texttt{Cooperative 3D Coverage} & \notag\\
%&\hspace*{-10mm}~~~~~~\texttt{3D Coverage Controller} & \notag\\
& \hspace*{-3mm}~~~~\underset{\{u^j_{k+\kappa|k},~\theta^j_{k+\kappa|k},~\phi^j_{k+\kappa|k}\}_{j=1}^{N}}{\arg \min} ~\mathcal{F}_\text{mission}, &  \hspace*{-20mm} \label{eq:objective_P2} \\
&\hspace*{-3mm}\textbf{subject to: $\kappa \in [1,..,K]$} ~  &\nonumber\\
&\hspace*{-3mm} x^j_{k+\kappa|k} = A x^j_{k+\kappa-1|k} + B u^j_{k+\kappa|k} & \hspace*{-3mm} \forall \kappa, j \label{eq:P2_1}\\
&\hspace*{-3mm} x^j_{k|k} = x^j_{k|k-1} & \hspace*{-3mm} \forall \kappa, j\label{eq:P2_2}\\
&\hspace*{-3mm} \mathcal{V}^j_{\hat{\xi},k+\kappa|k} =  \tilde{\mathcal{V}}^j_{\hat{\xi},k+\kappa|k} + x_{k+\kappa|k}^{j,\mathbf{p}} & \hspace*{-30mm} \forall \kappa, \hat{\xi}, j \label{eq:P2_3}\\
&\hspace*{-3mm} \sum_{\hat{\xi}=[1,..,|\Xi|]} \nu^j_{\hat{\xi},k+\kappa|k} = 1 & \hspace*{-3mm} \forall \kappa, j \label{eq:P2_4}\\
&\hspace*{-3mm} b^{j}_{\hat{\tau},\hat{\xi},k+\kappa|k} = 1 \iff  \tau \in  \bigtriangleup \left(\mathcal{V}^j_{\hat{\xi},k+\kappa|k}\right) & \hspace*{-3mm} \forall \hat{\tau},  \kappa,  \hat{\xi},  j \label{eq:P2_5}\\
&\hspace*{-3mm} \bar{b}^j_{\tau,\xi,k+\kappa|k} = \nu^j_{\hat{\xi},k+\kappa|k}~\wedge~ & \hspace*{-3mm} \forall \hat{\tau}, \kappa, \hat{\xi}, \hat{c}, j \label{eq:P2_6}\\
&\hspace*{+15mm} (b^{j}_{\hat{\tau},\hat{\xi},k+\kappa|k}\wedge~ \rho_{\hat{c},{\hat{\tau}}}\wedge \tilde{\rho}^j_{\hat{c}}) & \hspace*{-3mm} \notag\\
&\hspace*{-3mm} \hat{b}^j_{\hat{\tau},\hat{\xi},k+\kappa|k} \leq \bar{b}^j_{\hat{\tau},\hat{\xi},k+\kappa|k} + \mathcal{Q}(\tau) & \hspace*{-3mm} \forall \hat{\tau}, \kappa, \hat{\xi}, j \label{eq:P2_7}\\
&\hspace*{-3mm} \sum_{j} \sum_{\hat{\xi}} \sum_{\kappa} \hat{b}^j_{\hat{\tau},\hat{\xi},k+\kappa|k} \leq 1 & \hspace*{-3mm} \forall \hat{\tau} \label{eq:P2_8}\\
%&\hspace*{-3mm}  w^j_k = \sum_{\tau} \sum_{\xi} \sum_{\kappa} \hat{b}^j_{\tau,\xi,k+\kappa|k}  & \hspace*{-3mm} \forall j\label{eq:P2_9}\\
%&\hspace*{-3mm}  \left( \varepsilon^i_k = |w^{i_1}_k - w^{i_2}_k| \right) < \omega  & \hspace*{-3mm} \forall i, i_1, i_2, k \label{eq:P2_10}\\
&\hspace*{-3mm} x_{k+\kappa|k}^{j,\mathbf{p}} \notin \bigtriangleup(\psi) & \hspace*{-3mm} \forall \psi \in \Psi, \kappa, j\label{eq:P2_9}\\
&\hspace*{-3mm} x_{k+\kappa|k}^{j,\mathbf{p}} \notin \bigcup_{i \ne j = 1}^{N} \bigtriangleup(\mathcal{A}^i) & \hspace*{-3mm} \forall \kappa, j\label{eq:P2_10}
%&\hspace*{-3mm} x^j_{k+\kappa|k} \in \mathcal{X},~ u^j_{k+\kappa|k} \in \mathcal{U} & \hspace*{-3mm} \notag\\ 
%&\hspace*{-3mm}  \nu^j_{\xi,k+\kappa|k},~  b^{\hat{\mathcal{V}}_j}_{\tau,\xi,k+\kappa|k},~ \bar{b}^j_{\tau,\xi,k+\kappa|k}, \in \{0,1\}   &  \hspace*{-3mm} \notag\\ 
%&\hspace*{-3mm}  \hat{b}^j_{\tau,\xi,k+\kappa|k},~ \mathcal{Q}(\kappa) \in \{0,1\}, \xi \in [1,..,|\Xi|]   & \hspace*{-3mm}  \notag\\ 
%&\hspace*{-3mm} \tau\in [1,..,|\mathcal{T}|], i \in [1,..,|\mathcal{C}^N_2|],  (i_1,i_2) \in \mathcal{C}^N_2 & \hspace*{-3mm} \notag
\end{align}
\end{subequations}
\vspace{-8mm}
\end{algorithm}

\subsection{Constraints} \label{ssec:constraints}

The generation of look-ahead trajectories is achieved with the constraints shown in Eq. \eqref{eq:P2_1}-\eqref{eq:P2_2}, by appropriately selecting the control inputs $u^j_{k+\kappa|k}$ inside the horizon for all agents, according to the kinematic model as discussed in Sec. \ref{ssec:kinematic_model}.
The constraint shown in Eq. \eqref{eq:P2_3} rotates and translates agent's $j$ camera FOV inside the planning horizon. More specifically, $\mathcal{V}^j_{\hat{\xi},k+\kappa|k}$ denotes the $\hat{\xi}_\text{th} \in [1,..,|\Xi|]$ configuration of the camera's FOV vertices at the future time-step $\kappa$. The set of all possible $|\Xi|$ FOV rotations is precomputed as $\tilde{\mathcal{V}}^{j,i}_{\hat{\xi},k+\kappa|k} =  R_z(\phi^j_{k+\kappa|k})R_y(\theta^j_{k+\kappa|k})\mathcal{V}^i_0, \forall i \in [1,..,5], \forall \hat{\xi} \in [1,..,|\Xi|]$, where $(\theta^j_{k+\kappa|k} \in \Theta,\phi^j_{k+\kappa|k} \in \Phi) \in \Xi$, and then translated to the agent's position $x_{k+\kappa|k}^{j,\mathbf{p}}$ as shown in Eq. \eqref{eq:P2_3}.
Next, the binary variable $\nu^j_{\hat{\xi},k+\kappa|k} \in \{0,1\}$ indicates which of the $|\Xi|$ camera FOV configurations is active at time-step $\kappa$, to avoid double counting the same facet with different FOV configurations. This is achieved with the constraint shown in Eq. \eqref{eq:P2_4}.

To determine whether facet $\tau$ resides inside the agent's $j$, $\hat{\xi}_\text{th}$ camera FOV configuration at time-step $\kappa$ i.e., $\tau \in  \bigtriangleup(\mathcal{V}^j_{\hat{\xi},k+\kappa|k})$ first observe that an arbitrary point $p \in \mathcal{E}$ which belongs to the convex-hull defined by the camera FOV vertices $\mathcal{V}$ satisfies the following system of linear inequalities: $\alpha_{n} \cdot p \leq \beta_{n}, ~ \forall n=[1,..,5]$, where $\alpha_{n} \cdot p = \beta_{n}$ is the equation of the plane which contains the $n_\text{th}$ face of the FOV (with 5 faces in total), $\alpha_{n}$ is the unit outward normal vector to the plane containing the $n_\text{th}$ FOV face, and $\beta_{n}$ is a constant. Any point $p \in \mathcal{E}$ which satisfies the aforementioned system of inequalities is contained within the convex-hull of $\mathcal{V}$, and therefore can be potentially observed by the agent (provided it is visible). Subsequently, the binary variable $b^{j}_{\hat{\tau},\hat{\xi},k+\kappa|k} \in \{0,1\}$ shown in Eq. \eqref{eq:P2_5} is activated when facet $\tau$ resides inside the $\hat{\xi}_\text{th}$ camera FOV configuration of agent $j$ at time-step $\kappa$. This functionality can be defined as shown below:
\begin{subequations}
\begin{align} 
&  \alpha^j_{n,\hat{\xi},\kappa} \cdot \tau + o^{j}_{n,\hat{\tau},\hat{\xi},\kappa}(M-\beta^j_{n,\hat{\xi},\kappa}) \le M,~ \forall n,\hat{\tau}, \hat{\xi}, \kappa, \label{eq:chull1}\\
& 5b^{j}_{\hat{\tau},\hat{\xi},\kappa} - \sum_{n=1}^5 o^{j}_{n,\hat{\tau},\hat{\xi},\kappa} \le 0, ~ \forall \kappa, \hat{\xi}, \hat{\tau}. \label{eq:chull2}
\end{align}
\end{subequations}

\begin{figure*}
	\centering
	\includegraphics[width=\textwidth]{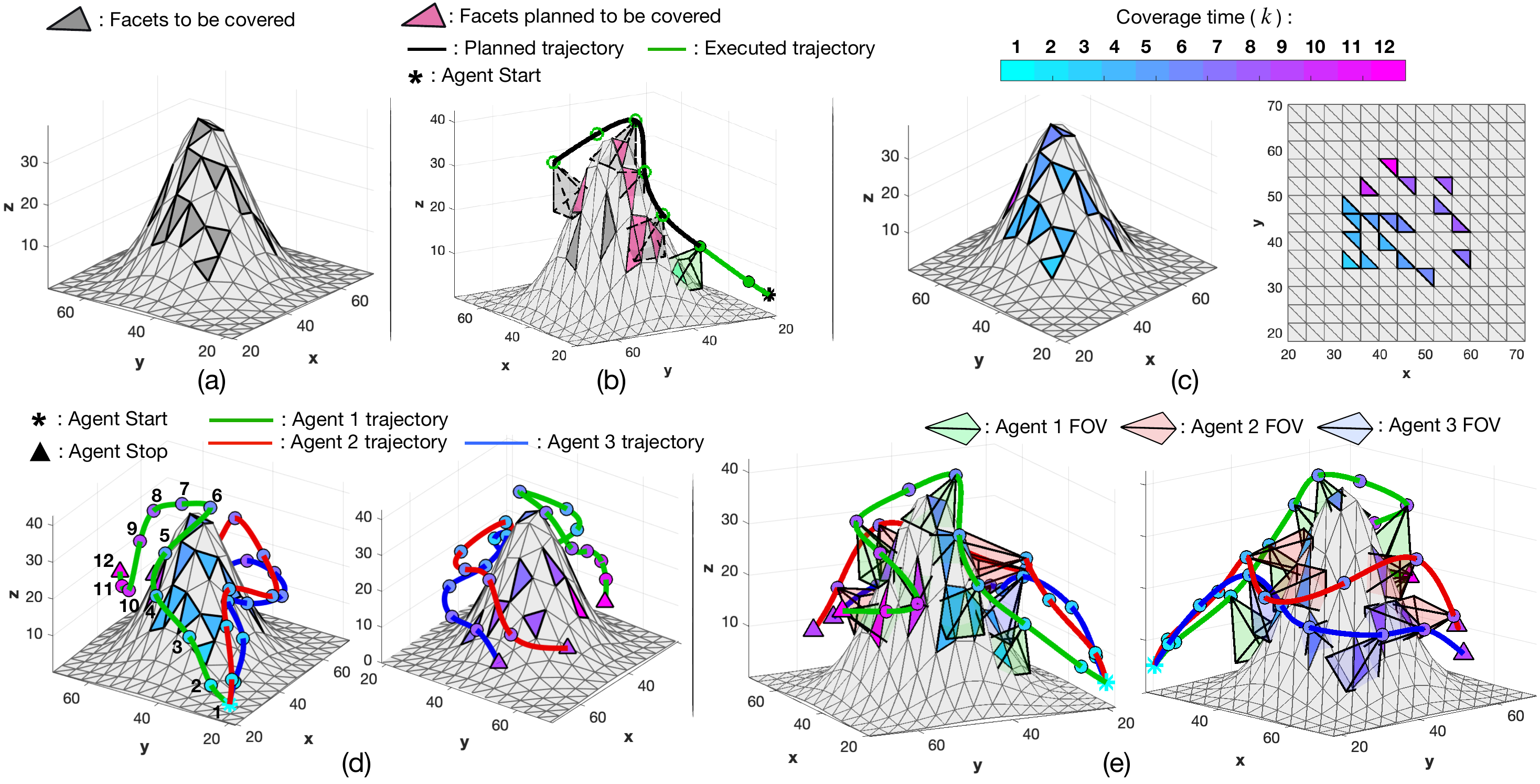}
	\caption{The figure shows an illustrative example of the proposed cooperative receding horizon 3D coverage controller with a team of 3 agents.}
	\label{fig:res1}
	\vspace{-5mm}
\end{figure*}

\noindent where $k+\kappa|k$ is abbreviated as $\kappa$, $\alpha^j_{n,\hat{\xi},\kappa}$ and $\beta^j_{n,\hat{\xi},\kappa}$ are the plane equation coefficients, containing the $n_\text{th}$ FOV face of the $\hat{\xi}_\text{th}$ FOV configuration, of the $j$ agent at time-step $\kappa$, $o^{j}_{n,\hat{\tau},\hat{\xi},\kappa} \in \{0,1\}$ is an auxiliary binary variable which is activated when the $n_\text{th}$ inequality is true i.e., $\alpha^j_{n,\hat{\xi},\kappa} \cdot \tau \leq \beta^j_{n,\hat{\xi},\kappa}$,  and $M$ is a large positive constant that ensures the constraint is valid when $o^{j}_{n,\hat{\tau},\hat{\xi},\kappa}=0$. Finally, $b^{j}_{\hat{\tau},\hat{\xi},\kappa}$ is activated to satisfy Eq. \eqref{eq:chull2} when $o^{j}_{n,\hat{\tau},\hat{\xi},\kappa}=1, \forall n$ thus indicating that facet $\tau$ resides inside the $\hat{\xi}_\text{th}$ camera FOV configuration of agent $j$. For brevity facet $\tau$ is represented by its centroid in this formulation. 

 The logical conjunction shown in Eq. \eqref{eq:P2_6} makes sure that agent $j$ does not activates more than one camera FOV configurations at each time-step $\kappa$, with the binary variable $\nu^j_{\hat{\xi},k+\kappa|k}$. The constraint in Eq. \eqref{eq:P2_6} also checks whether the facet $\tau$ which resides inside the agent's FOV is visible via the learned visibility determination variable $\rho_{\hat{c},\hat{\tau}}$. 
% As discussed in Sec. \ref{ssec:visibility_determination} $\rho_{\hat{c},\hat{\tau}}$ is activated when there exists a FOV pose which results in the visibility of facet $\tau$ when the agent resides inside cell $c$. 
Subsequently, the logical conjunction $\nu^j_{\hat{\xi},k+\kappa|k}\wedge (b^{j}_{\tau,\xi,k+\kappa|k}\wedge~ \rho_{\hat{c},{\hat{\tau}}}\wedge \tilde{\rho}^j_{\hat{c}})$ becomes true when at time-step $\kappa$ facet $\tau$ resides within the convex-hull of the active FOV configuration $\xi$ (as indicated by the variables $\nu^j_{\hat{\xi},k+\kappa|k}$ and $b^{j}_{\tau,\xi,k+\kappa|k}$), and at the same time the agent position $x_{k+\kappa|k}^{j,\mathbf{p}}$ resides within the cell $c$ (as indicated by the binary variable $\tilde{\rho}^j_{\hat{c}}$) from which it has been determined that the facet $\tau$ is visible via the learned visibility variable $\rho_{\hat{c},{\hat{\tau}}}$. 
%This result is then saved to the binary variable $\bar{b}^{j}_{\hat{\tau},\hat{\xi},k+\kappa|k} \in \{0,1\}$ which essentially indicates the coverage of facet $\tau$, at time-step $\kappa$, with the FOV configuration $\xi$ and the agent state $x^j_{k+\kappa|k}$. The constraint $\tilde{\rho}^j_{\hat{c}}=1 \iff x_{k+\kappa|k}^{j,\mathbf{p}} \in \bigtriangleup(c)$ is implemented similarly to the constraints shown in Eq. \eqref{eq:chull1}-\eqref{eq:chull2}.
The constraint shown in Eq. \eqref{eq:P2_7} is used for avoiding the duplication of work (i.e., avoiding to cover facets that have been covered in the past). To achieve this, the function $\mathcal{Q}: \mathcal{T} \rightarrow \{0, 1\}$ keeps track of all the facets that have been covered by the agents up to the current time-step $k$. Therefore any facet $\tau$ that has been covered by any agent $j$ results in $\mathcal{Q}(\tau) = 1$. Consequently, the binary variable $\hat{b}^{j}_{\hat{\tau},\hat{\xi},k+\kappa|k}$ in Eq. \eqref{eq:P2_7} is maximized for facet $\tau$ either through $\mathcal{Q}(\tau)$ or via $\bar{b}^{j}_{\hat{\tau},\hat{\xi},k+\kappa|k}$. For this reason, the agent $j$ has no incentive to plan a coverage trajectory for facet $\tau$ inside the planning horizon, when $\mathcal{Q}(\tau) = 1$ since the binary variable  $\hat{b}^{j}_{\hat{\tau},\hat{\xi},k+\kappa|k}$ is maximised through $\mathcal{Q}(\tau)$.

The constraint shown in Eq. \eqref{eq:P2_8} makes sure that during the planning horizon the facet $\tau$ is not planned to be covered more than once by the same or multiple agents, and then the constraints shown in Eq. \eqref{eq:P2_9} ensure that the agents avoid collisions with the obstacles $\psi \in \Psi$ in the environment, including the object of interest. This is achieved by enforcing the agent position $x_{k+\kappa|k}^{j,\mathbf{p}}$ to reside outside the convex-hull of all obstacles $\psi \in \Psi$ at all time-steps $\kappa$. Assuming that the convex-hull of obstacle $\psi$, which is represented as a triangle mesh $\mathcal{T}_{\hat{\psi}}$ (where $\hat{\psi} \in [1,..,|\Psi|]$ is the index of $\psi$) is given by the intersection of $n_{\hat{\psi}}$ half-spaces, where the $i_\text{th}$ half-space is associated with the plane equation $\alpha^{{\hat{\psi}}}_{i} \cdot p = \beta^{{\hat{\psi}}}_{i},~ i\in [1,..,n_{\hat{\psi}}],  p \in \mathcal{E}$, which divides the 3D space into two parts, the obstacle avoidance constraints for all obstacles can be defined as follows:
\begin{align}
&  \alpha^{{\hat{\psi}}}_{i} \cdot x_{k+\kappa|k}^{j,\mathbf{p}} + M z^{j,{\hat{\psi}}}_{k+\kappa|k,i} > \beta^{{\hat{\psi}}}_{i},~\forall j, \kappa, i, {\hat{\psi}}, \label{eq:O_1}\\
& \sum_{i=1}^{n_{\hat{\psi}}} z^{j,{\hat{\psi}}}_{k+\kappa|k,i} \le  (n_{\hat{\psi}}-1), ~ \forall j, \kappa, {\hat{\psi}} \label{eq:O_2}
\end{align}
\noindent where $z^{j,{\hat{\psi}}}_{k+\kappa|k,i} \in \{0,1\}$ is a binary variable which indicates when activated that $\alpha^{{\hat{\psi}}}_{i} \cdot x_{k+\kappa|k}^{j,\mathbf{p}} > \beta_i^{{\hat{\psi}}}$ is not true. Therefore when $z^{j,{\hat{\psi}}}_{k+\kappa|k,i} = 1, \forall i \in [1,..,n_{\hat{\psi}}]$ indicates that agent $j$ resides within the convex-hull of obstacle $\psi$ at time-step $\kappa$. Consequently, a collision is avoided with obstacle $\psi$ at time-step $\kappa$ when $\exists i \in [1,..,n_{\hat{\psi}}] : z^{j,{\hat{\psi}}}_{k+\kappa|k,i} = 0$ which is achieved with the constraint in Eq. \eqref{eq:O_2}. The same principle is applied to implement collision avoidance constraints amongst the team of agents as shown in Eq. \eqref{eq:P2_10}, which requires that during all time-steps the agent's $j$ positional state must reside outside the convex-hull of agent's $i$ safety area $\mathcal{A}^i,i \ne j, i \in [1,..,N]$, where $\mathcal{A}^i$ is the inscribed dodecahedron around $x_{k+\kappa|k}^{i,\mathbf{p}}$ which approximates a spherical safety area with certain radius around the agent \cite{PapaioannouTMC2022}. Finally, the  mission is terminated when  $\sum_{\tau \in \mathcal{T}} \mathcal{Q}(\tau) = |\mathcal{T}|$.

%% file: evaluation.tex
\section{Evaluation} \label{sec:Evaluation}
\subsection{Simulation Setup} \label{ssec:sim_setup}

%\begin{figure}
%	\centering
%	\includegraphics[width=\columnwidth]{figs/res2.pdf}
%	\caption{The  figure  shows  how  the  object's  surface  area  has been  split  amongst the agents during the coverage mission.}
%	\label{fig:res2}
%	\vspace{-2mm}
%\end{figure}
%
%\begin{figure}
%	\centering
%	\includegraphics[width=\columnwidth]{figs/res3.pdf}
%	\caption{The figure illustrates the coverage performance of the proposed approach in terms of the mission elapsed time.}
%	\label{fig:res3}
%	\vspace{-6mm}
%\end{figure}
%
%\begin{figure*}
%	\centering
%	\includegraphics[width=\textwidth]{figs/res4-compressed.pdf}
%	\caption{Experimental evaluation of the proposed cooperative 3D coverage controller.}
%	\label{fig:res4}
%	\vspace{-7mm}
%\end{figure*}

For the evaluation of the proposed approach we assume agents with identical capabilities. Subsequently, the agent's $j$ kinematic model parameters $\Delta T$, $\gamma$, and $m$ are set to 1s, 0.2, and 1.05kg respectively. The agent velocity $x_k^{j,\mathbf{v}}$ is bounded within the interval $[-12,12]$m/s, whereas the kinematic control input $u^j_k$ is bounded within the interval $[-10,10]$N. The agent camera FOV model parameters $(\ell, w, r)$ are set to $(10, 10, 16)$m, and the gimbal rotation angles $\theta$ and $\phi$ take their values from the finite sets $\Theta=\{30, 90, 150\}$deg, and  $\Phi=\{30, 105, 180, 255, 330\}$ respectively, leading to $|\Xi|=15$ possible camera FOV configurations. The 3D environment $\mathcal{E}$ is bounded in each dimension in the interval $[0,100]$m, and the object of interest to be covered is given by the Gaussian function $f(x,y) = A\exp\left(-\left(\frac{(x-x_o)^2}{2\sigma^2_x}+\frac{(y-y_o)^2}{2\sigma^2_y}\right)\right)$, with $(x_o,y_o) = (45, 45)$, $\sigma^2_x=\sigma^2_y = 80$, and $A = 40$, which has been Delaunay triangulated into  $|\mathcal{T}| = 220$ triangular facets. For the visibility determination constraints we have used 50 light rays i.e., $|\mathcal{L}^j_k(\theta^j_k,\phi^j_k,x_k^{j,\mathbf{p}})| = 50$, $N_r$ was set to 100, and the procedure described in Sec. \ref{ssec:visibility_determination} was conducted on a discretized version of the environment $\tilde{\mathcal{E}}$, composed of $|\tilde{\mathcal{E}}|=1000$ non-overlapping 3D cuboid cells. The planning horizon in the following experiments has been set to $K=5$. % and finally Problem (P2) has been implemented and executed with Gurobi solver.

\vspace{-0mm}
\subsection{Simulation Experiment}
An illustrative example of the proposed approach with 3 UAV agents is shown in Fig. \ref{fig:res1}. Specifically, Fig. \ref{fig:res1}(a) shows the object of interest to be covered. Without loss of generality, and in order to aid the analysis and visual clarity of this demonstration, we only require that a random subset of facets $\tilde{\mathcal{T}} \subseteq \mathcal{T}$ needs be covered by the agents, instead of the full triangle mesh  $\mathcal{T}$, thus we randomly sample 18 facets ($|\tilde{\mathcal{T}}|=18$) as shown in Fig. \ref{fig:res1}(a) marked with dark gray color. Figure \ref{fig:res1}(b) shows the controller's output for agent 1 at time-step $k=3$, with the executed trajectory shown in green color, and the predicted trajectory shown in black. The figure illustrates the generated finite-length look-ahead trajectory (i.e., kinematic and camera states) of agent 1 inside the planning horizon $k+\kappa|k,~k=3, \kappa=[1,..5]$. Then Fig. \ref{fig:res1}(c) shows the time-steps at which the facets $\tau \in \tilde{\mathcal{T}}$ have been covered by the agents (both in 3D and top-down view), color-coded based on the coverage time. Figure \ref{fig:res1}(d) shows in detail (front and back view of the object of interest) the kinematic trajectories of the 3 agents during the coverage mission, indicating the time-steps at which the agents cover the object's facets with different color. Finally, Fig. \ref{fig:res1}(e)  shows the camera FOV configurations of each agent used during the coverage mission.

%% file: conclusion.tex
\vspace{-1mm}
\section{Conclusion} \label{sec:conclusion}
We have proposed a cooperative coverage controller for 3D environments which allows a team of networked UAV agents to work cooperatively in order to cover the total surface area of an object of interest. We have formulated the coverage planning problem as a receding horizon optimal control problem which jointly optimizes the kinematic and camera control inputs over all agents, under duplication of work constraints and visibility determination constraints. Future work will investigate the extension of the proposed approach to a distributed system, and study how uncertainty can be handled using robust control techniques.